\documentclass[11pt, a4paper, logo, copyright, nonumbering]{deepseek}
\usepackage[numbers, sort&compress, round]{natbib}
\usepackage{dblfloatfix}
\usepackage{ulem}
\usepackage{caption}
\usepackage{listings}
\usepackage{verbatim}
\usepackage{dramatist}
\usepackage{xspace}
\usepackage{pifont}
\usepackage{multirow}
\usepackage{xltabular}
\usepackage{longtable}
\usepackage{hyperref}
\usepackage{tcolorbox}
\usepackage{tikz}
\usepackage{fontawesome5} 
\tcbuselibrary{breakable} 
\tcbuselibrary{skins,listings} 
\usepackage{amsfonts}
\usepackage{amsmath}
\usepackage{xcolor} 
\usepackage{amssymb}
\usepackage{lineno}
\usepackage{adjustbox}
\usepackage{graphicx}
\usepackage[bottom]{footmisc}

\usepackage{CJKutf8}
\usepackage{subcaption}
\usepackage{setspace}

\usepackage{dsfont}
\usepackage{array}
\usepackage{tabularx}
\usepackage{booktabs}
\usepackage{lipsum}
\usepackage{multicol}
\definecolor{promptgray}{gray}{0.9} 
\makeatletter
\def\@BTrule[#1]{%
  \ifx\longtable\undefined
    \let\@BTswitch\@BTnormal
  \else\ifx\hline\LT@hline
    \nobreak
    \let\@BTswitch\@BLTrule
  \else
     \let\@BTswitch\@BTnormal
  \fi\fi
  \global\@thisrulewidth=#1\relax
  \ifnum\@thisruleclass=\tw@\vskip\@aboverulesep\else
  \ifnum\@lastruleclass=\z@\vskip\@aboverulesep\else
  \ifnum\@lastruleclass=\@ne\vskip\doublerulesep\fi\fi\fi
  \@BTswitch}
\makeatother

\addto\extrasenglish{

}

\reportnumber{001}

\renewcommand{\today}{}

\title{\centering HiDream-O1-Image: A Natively Unified Image Generative Foundation Model with Pixel-level Unified Transformer}

\author[*]{
\vspace{-0.2in}
HiDream.ai
\vspace{-0.2in}
}

\renewcommand{\phi}{\varphi}

\renewcommand{\epsilon}{\varepsilon}
\renewcommand{\imath}{\mathrm{i}}

\newlength{\restsubwidth}
\newlength{\restsubheight}
\newlength{\restsubmoreheight}
\newcommand{\rest}[2]{%
        \settowidth{\restsubwidth}{\ensuremath{#2}}
        \settoheight{\restsubheight}{\ensuremath{{}_{#2}}}
        \ensuremath{{#1\hskip 0.5pt}_{\vrule\kern2pt\parbox[b][%
        4pt][b]{\the\restsubwidth}{%
                        \ensuremath{{}_{#2}}}}}
        }

\vspace{-1in}

\begin{abstract}

The evolution of visual generative models has long been constrained by fragmented architectures relying on disjoint text encoders and external VAEs. In this report, we present \textbf{HiDream-O1-Image}, a natively unified generative foundation model via pixel-space Diffusion Transformer, that pioneers a paradigm shift from modular architectures to an end-to-end in-context visual generation engine. By mapping raw image pixels, text tokens, and task-specific conditions into a single shared token space, HiDream-O1-Image achieves a structural unification of multimodal inputs within an \textbf{Unified Transformer (UiT)} architecture. This native encoding paradigm eliminates the need for separate VAEs or disjoint pre-trained text encoders, allowing the model to treat diverse generation and editing tasks as a consistent in-context reasoning process. Extensive experiments show that HiDream-O1-Image excels across various generation tasks, including text-to-image generation, instruction-based editing, and subject-driven personalization. Notably, with only 8B parameters, \textbf{HiDream-O1-Image (8B)} achieves performance parity with or even surpasses established state-of-the-art models with significantly larger parameters (e.g., 27B Qwen-Image). Crucially, to validate the immense scalability of this paradigm, we successfully scale the architecture up to over 200B parameters. Experimental results demonstrate that this massive-scale version \textbf{HiDream-O1-Image-Pro (200B+)} unlocks unprecedented generative capabilities and superior performance, establishing new state-of-the-art benchmarks. Ultimately, HiDream-O1-Image highlights the immense potential of natively unified architectures and charts a highly scalable path toward next-generation multimodal AI.

\textbf{Github:} \url{https://github.com/HiDream-ai/HiDream-O1-Image}

\textbf{Huggingface:} \url{https://huggingface.co/HiDream-ai/HiDream-O1-Image}

  \end{abstract}
  
\begin{document}
  
  \maketitle
  
  \begin{figure}[h]
  \vspace{-0.1in}
  \centering
  \includegraphics[width=0.96\textwidth]{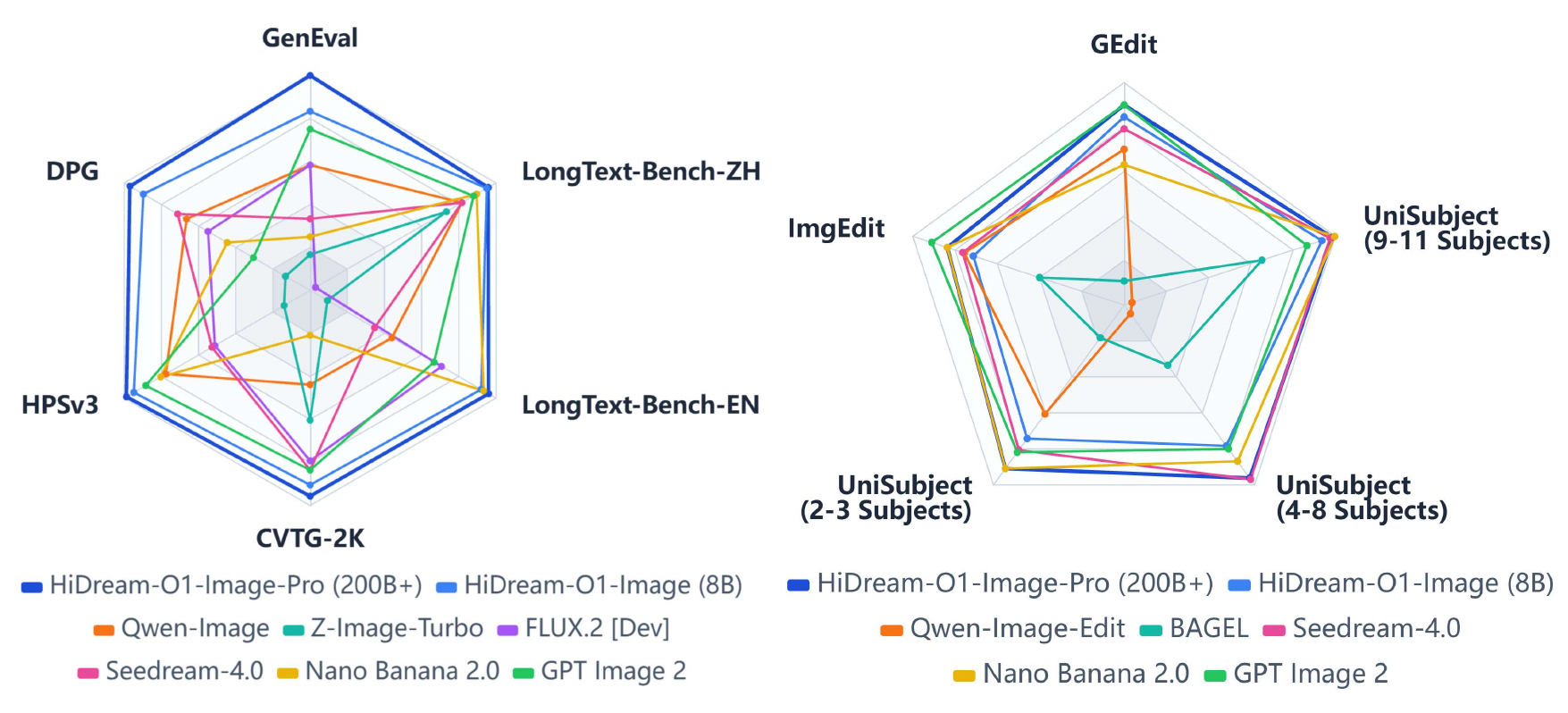}
    \vspace{-0.1in}
  \caption{
  \centering
  HiDream-O1-Image shows strong capabilities across various benchmarks and tasks.
  }
  \label{fig:dsv3_performance}
  \end{figure}
  
  \newpage
  
  \begin{spacing}{0.9}
  \tableofcontents
  \end{spacing}
  
    \begin{figure}
       \vspace{-0.2in}
    \centering
    \includegraphics[width=0.96\linewidth]{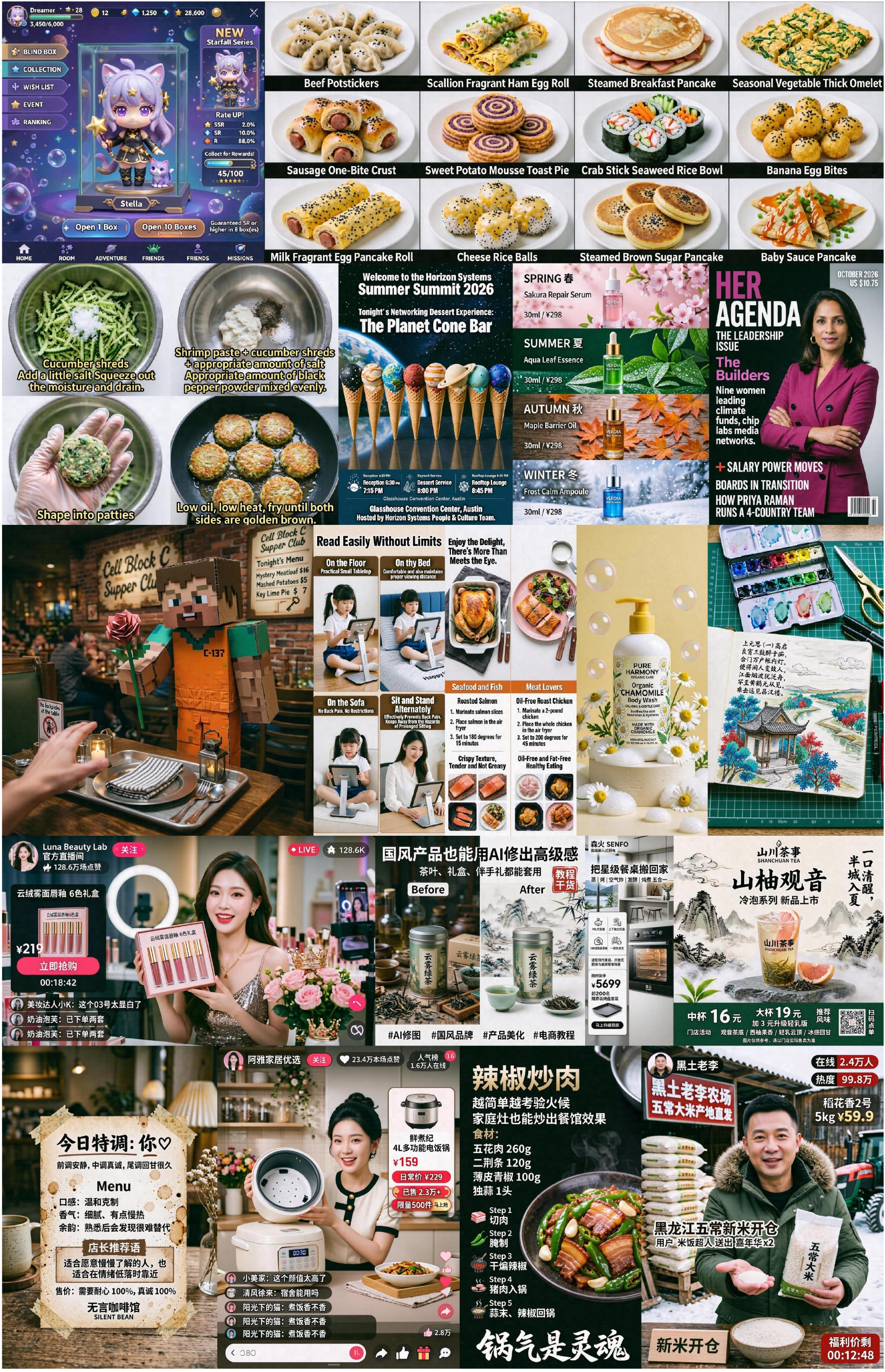} 
    \caption{Showcases of HiDream-O1-Image on text-to-image task with complex text rendering.}
    \label{fig:gen_images_ocr}
  \end{figure}

  \begin{figure}
    \vspace{-0.2in}
    \centering
    \includegraphics[width=1.0\linewidth]{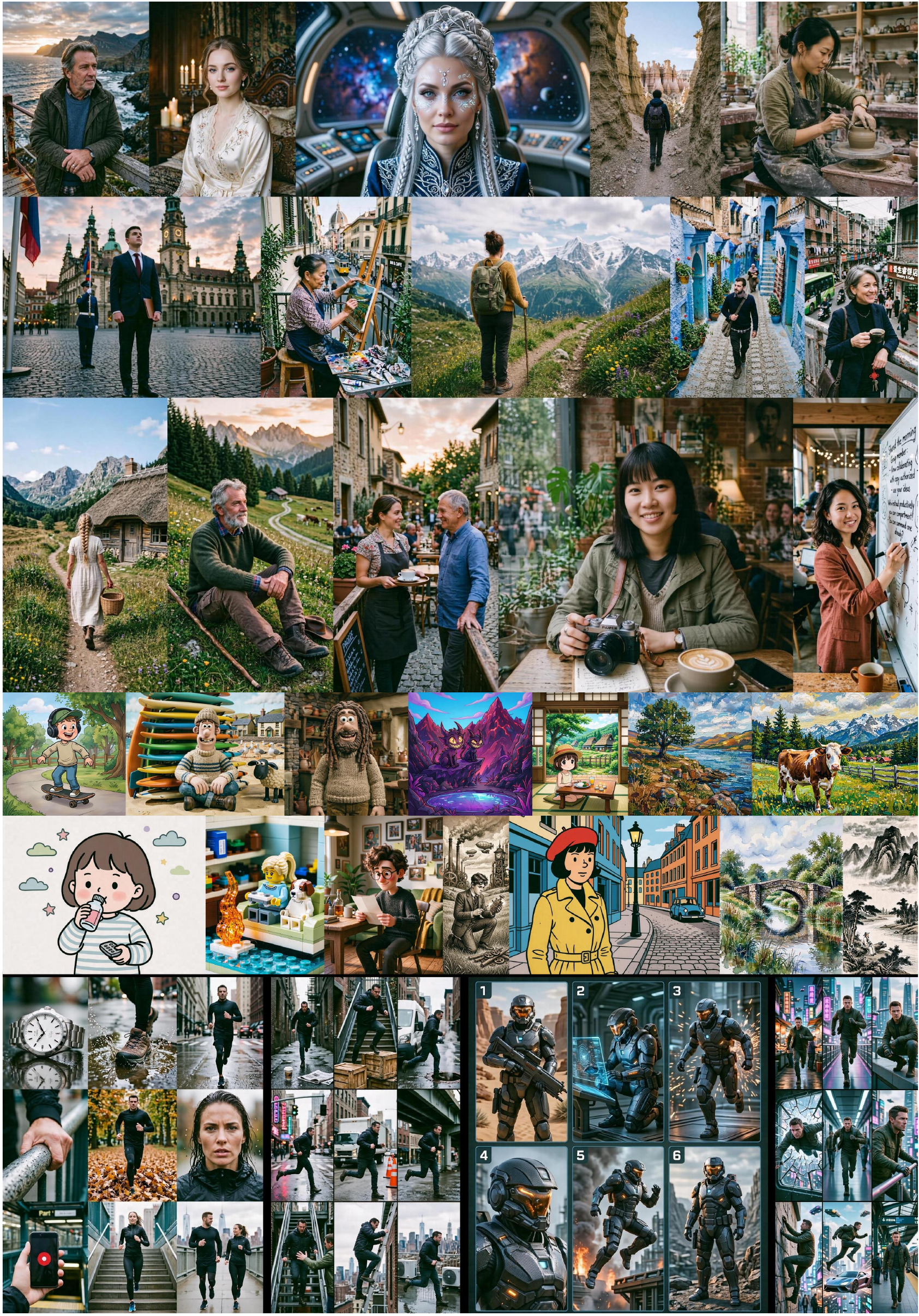} 
    \caption{Showcases of HiDream-O1-Image on text-to-image task in diverse cinematic shots, versatile artistic styles, and multi-panel image generation scenarios.}
    \label{fig:gen_images}
  \end{figure}

  \begin{figure}
    \vspace{-0.3in}
    \centering
    \includegraphics[width=1.0\linewidth]{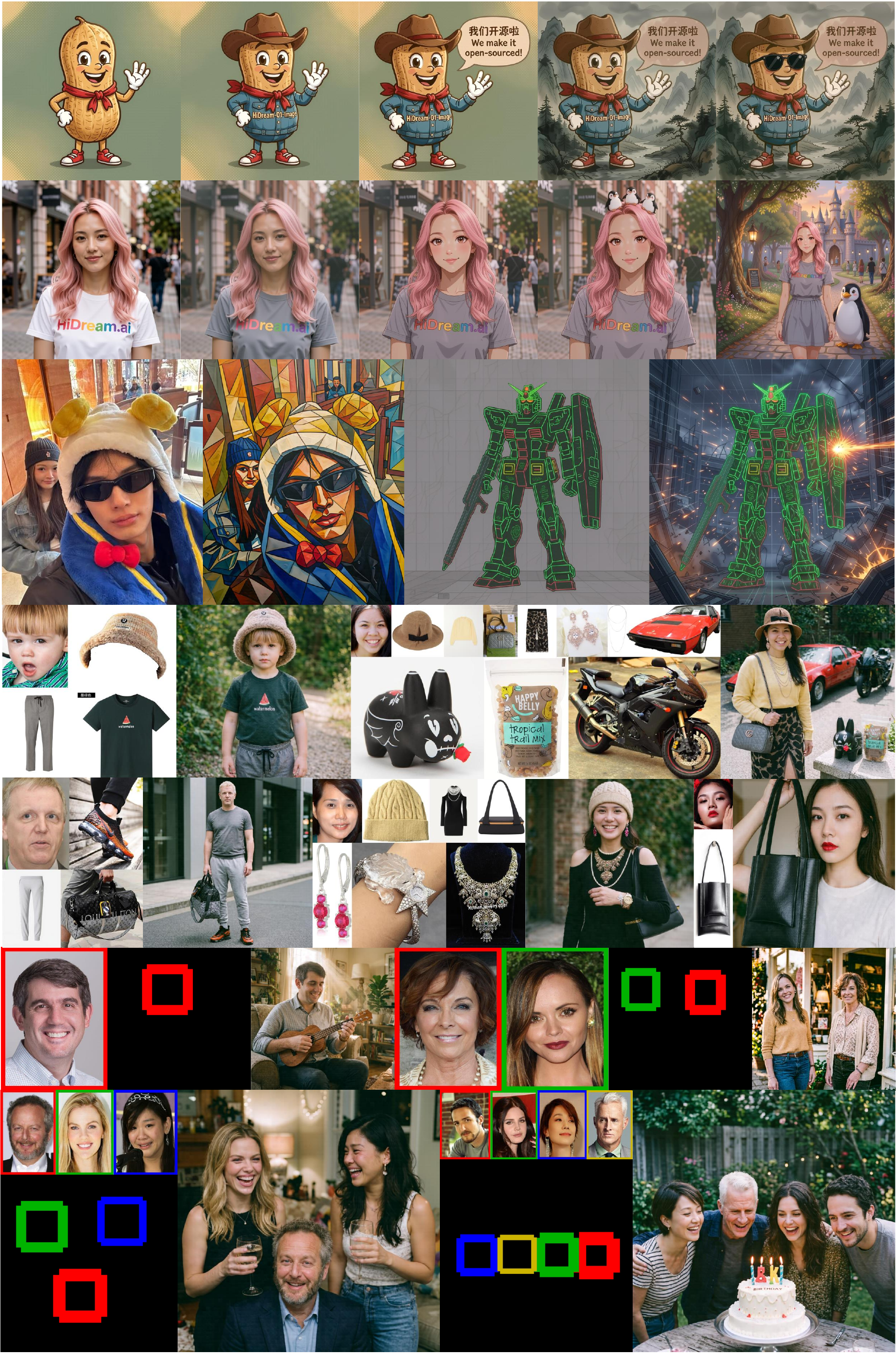}     \vspace{-0.3in}
    \caption{Showcases of HiDream-O1-Image on instruction-based editing and subject-driven personalization tasks.}
    \label{fig:gen_images}
  \end{figure}

\newpage
\section{Introduction}

\begin{figure*}
    \centering {\includegraphics[width=1\textwidth]{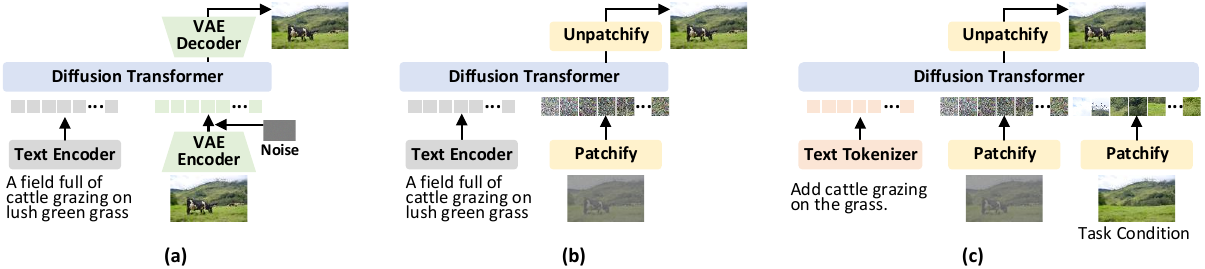}}
    \vspace{-0.0in}
    \caption{Unlike (a) \textbf{latent DiTs} that use latent-space VAE compression and (b) \textbf{pixel-space DiTs} that typically rely on disjoint text encoder, (c) \textbf{Unified Transformer} in our HiDream-O1-Image natively encodes raw image pixels, texts, and task-specific conditions within a shared token space, and thus generalizes to broader and more complex generative tasks.}
    \label{fig:intro}
    \vspace{-0.0in}
\end{figure*}

The landscape of visual content generation has been fundamentally reshaped by the rapid evolution of diffusion models \cite{ho2020denoising,flux1,sd3medium,ma2025janusflow,qwenimage,zheng2025hierarchical,yao2025denoising,mao2025visual}. Recently, the architecture of generative models has witnessed a significant transition from the traditional U-Net \cite{rombach2022ldm} to the Diffusion Transformer (DiT) \cite{dit,zhu2024sd}, pushing the boundaries of image and video synthesis \cite{xiao2025omnigen,bagle}. Amidst this progress, the dominant paradigm remains anchored in Latent Diffusion Models (LDMs) \cite{rombach2022ldm}. LDMs rely on a modular and fragmented pipeline: they utilize pre-trained Variational Autoencoders (VAEs) \cite{kingma2013vae} to compress raw images into a latent space, coupled with disjoint pre-trained language models (e.g., CLIP \cite{radford2021learning} or T5 \cite{raffel2020exploring}) to encode text prompts (Figure \ref{fig:intro} (a)). While computationally efficient, this disjoint encoding approach inevitably introduces information bottlenecks, e.g., the loss of high-frequency visual details during latent-space compression, thereby capping the upper bound of generation fidelity.

To bypass the structural limitations of latent-space compression, recent pioneering efforts have explored pixel-space Diffusion Transformers \cite{hoogeboom2025simpler,jit}. By modeling the diffusion process directly on raw image pixels, these approaches have demonstrated promising visual fidelity and intricate detail preservation in Text-to-Image (T2I) generation. However, despite discarding the VAE image encoder, most existing pixel-space DiTs (Figure \ref{fig:intro} (b)) still heavily rely on disjoint and off-the-shelf text encoders. This segregation of visual and textual encoding spaces inherently suffers from semantic misalignment, as the modalities are not jointly optimized from the ground up. Furthermore, these models typically remain specialized for single-task synthesis (primarily T2I), struggling to generalize to broader, more complex scenarios such as instruction-based image editing and subject-driven personalization.

In the realm of Natural Language Processing, the unification of diverse tasks into a single shared token space has paved the way for Large Language Models (LLMs) capable of powerful in-context reasoning. This success naturally motivates a critical question for visual generative foundation models: \textit{Can we scale a pixel-space diffusion model from a specialized generator into a versatile, generalist reasoning framework?} To achieve this, we must dismantle the boundaries between disparate encoding modules and structurally unify multimodal inputs at the foundational level, transitioning from modular pipelines to an end-to-end architecture.

In this work, we present \textbf{HiDream-O1-Image}, a natively unified generative foundation model driven by a new Pixel-level Unified Transformer. HiDream-O1-Image completely abandons the traditional fragmented encoding paradigm. Instead of relying on separate VAEs or disjoint pre-trained text encoders, our model maps raw image pixels, discrete text tokens, and auxiliary task-specific conditions directly into a single, continuous shared token space (Figure \ref{fig:intro} (c)). This structural unification allows all multi-modal inputs to be processed synergistically within such Unified Transformer architecture in an end-to-end fashion. By doing so, this native encoding paradigm empowers HiDream-O1-Image to treat diverse generation and editing tasks not as isolated problems requiring specialized modules, but as a consistent in-context visual reasoning process, fostering deeper and more flexible multi-modal interaction among inputs.

While architectural unification provides a powerful engine for visual reasoning, translating highly complex and abstract user intentions into model-preferred inputs remains a practical challenge. To bridge this semantic gap, we further introduce a \textbf{Reasoning-Driven Prompt Agent} equipped with a ``thinking'' mechanism. This agent explicitly reasons through and refines complex user instructions before feeding them into the generation pipeline. This mechanism significantly enhances the model's generalizability and instruction-following capabilities, particularly for intricate visual generation tasks that require deep logical deduction.

To the best of our knowledge, HiDream-O1-Image is among the first efforts to explore Pixel-level Unified Transformer architecture that simultaneously supports (i) various multi-modal primitive inputs (image, text, the sequence of additional conditions/references) and (ii) a unified system for text-to-image generation, instruction-based editing, and subject-driven personalization, while scaling effectively to high-resolution 2,048 $\times$ 2,048 outputs.

In summary, the main contributions of this work are highlighted as follows:
\begin{itemize}
\item \textbf{Natively Unified Generative Architecture:} We propose HiDream-O1-Image, an end-to-end Pixel-level Unified Transformer that completely discards traditional modular pipelines (i.e., external VAEs and disjoint text encoders). By mapping raw image pixels, text tokens, and task conditions into a single shared token space, we reframe diverse visual generation and editing tasks as a consistent in-context visual reasoning process.
\item \textbf{Reasoning-Driven Prompt Agent:} To bridge the semantic gap between raw user intentions and model-preferred inputs, we introduce and open-source a Prompt Agent equipped with a ``thinking'' mechanism. By explicitly reasoning through and refining complex user instructions, this agent significantly enhances the model's instruction-following capabilities, particularly for intricate, reasoning-heavy visual generation tasks.
\item \textbf{Exceptional Efficiency and Versatility at 8B Scale:} We demonstrate that our unified paradigm achieves state-of-the-art performance with high efficiency, unlocking comprehensive coverage across various generation scenarios. Specifically, HiDream-O1-Image seamlessly handles diverse cinematic shots, versatile artistic styles, complex long text rendering, instruction-based image editing, subject-driven personalization, and multi-panel image generation for storyboard production. Across these multifaceted scenarios, our model achieves parity with, or even surpasses, both established open-source latent-space DiTs with significantly larger parameters (e.g., the 27B Qwen-Image) and leading closed-source commercial models (e.g., Nano Banana 2.0).
\item \textbf{Immense Scalability to 200B+ Parameters:} We validate the scaling laws of our natively unified paradigm by successfully scaling HiDream-O1-Image architecture up to over 200B parameters. Experimental results at this massive scale unlock superior generative capabilities, visual fidelity, and intricate reasoning, establishing new state-of-the-art benchmarks across a wide spectrum of generation tasks.
\end{itemize}

\begin{figure}[ht]
  \centering
  \includegraphics[width=0.9\linewidth]{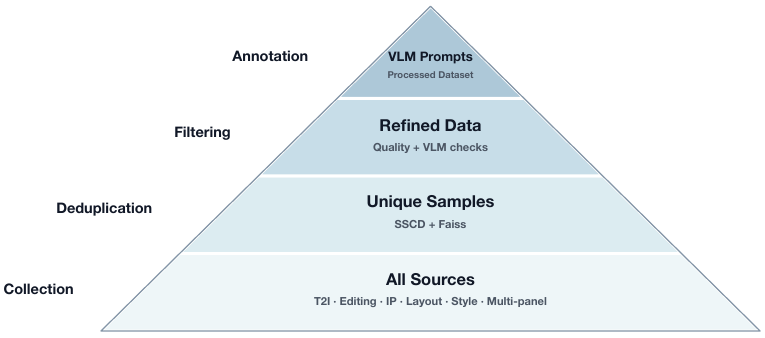} 
  \caption{Overview of data curation and prompt construction.}
  \label{fig:datapipeline}
\end{figure}

\section{Data Curation and Prompt Construction}

High-quality and large-scale training data is essential for scaling generalist image generation. Training a single model for text-to-image synthesis, instruction-based editing, and subject-driven personalization additionally requires supervision beyond standard image-text pairs. We therefore build a dedicated data engine that converts heterogeneous raw sources into high-quality image-text pairs, editing triplets, and subject-reference samples. As summarized in Figure~\ref{fig:datapipeline}, the pipeline consists of source data collection, data deduplication, data quality and safety filtering, and VLM-based prompt construction.

\noindent
\subsection{Source Data Collection}
We begin by assembling a large candidate pool from public web corpora and internally licensed collections. In addition to standard image-text pairs, we deliberately broaden the source distribution to cover all major training scenarios required by HiDream-O1-Image. For text-to-image synthesis, we include complex graphic-layout data such as presentation slides, posters, documents, and long-text images, which provide supervision for dense typography, structured composition, and mixed image-text layouts. We also increase the proportion of style-oriented samples, including photography styles, illustration styles, design templates, rendering aesthetics, and domain-specific visual identities.

Beyond text-to-image generation, we collect task-specific data for instruction-based editing and subject-driven personalization. The editing training data is constructed from public editing datasets, internally synthesized before-after pairs, and video-derived samples, where different frames provide natural supervision for object changes, background transitions, action variation, and local attribute modifications. For IP-oriented personalization, we gather both human-centered and object-centered reference sets. Human IP data contains multiple photographs of the same person across poses, expressions, viewpoints, lighting conditions, and scenes, while object IP data covers repeated appearances of the same item under varying backgrounds and configurations. Finally, we collect multi-panel data from two complementary sources: grid images crawled from the Internet and frame-composition samples constructed from different clips of the same video. These multi-panel data expose the model to sequential changes, panel-wise consistency, and richer spatial organization beyond single-image composition.

\noindent
\subsection{Data Deduplication}

Since large web-scale collections inevitably contain repeated or highly similar images, we apply a deduplication procedure to improve training efficiency and reduce memorization risk \cite{somepalli2023diffusion}. Direct pairwise comparison over the full corpus is infeasible at this scale, so we perform deduplication in two steps:
\begin{enumerate}
    \item \textit{Visual Feature Grouping.} We extract image representations using the SSCD model \cite{pizzi2022self}. A representative subset of 2 million samples is then used to fit k-means centroids, partitioning the feature space into 16,000 clusters so that likely duplicates are routed into the same local search space.
    \item \textit{Cluster-Level Similarity Search.} Within each cluster, we conduct nearest-neighbor retrieval with GPU-accelerated Faiss \cite{douze2024faiss}. Samples whose similarity scores exceed a predefined threshold are treated as near-duplicates, and only one representative image is retained.
\end{enumerate}
This redundancy-control stage removes approximately 20\% of the initial candidates while preserving the semantic coverage of the collected data.

\noindent
\subsection{Data Quality and Safety Filtering}
After deduplication, we further filter the remaining data with a set of complementary models. The goal is not only to remove harmful or low-quality images, but also to keep training samples that are visually informative and useful for high-resolution pixel-space modeling:
\begin{itemize}
    \item \textit{Safety Assessment.} Potentially inappropriate images are detected and removed by a pre-trained NSFW classifier \cite{laion2024clip}.
    \item \textit{Aesthetic Assessment.} We use an aesthetic scoring model \cite{laion2024aesthetic} to suppress images with poor visual appeal, while maintaining stylistic diversity across realistic, artistic, and design-oriented domains.
    \item \textit{Watermark Detection.} Images containing conspicuous watermarks are filtered by a dedicated watermark detector \cite{laion2024watermark}.
    \item \textit{Task Consistency Assessment.} For editing and IP-oriented data, we further employ a VLM to verify whether the samples form valid task instances. For editing data, the VLM judges whether the source and target images constitute a meaningful before-after pair with an interpretable visual change. For IP-related data, it checks whether the reference images correspond to the same person or the same object, filtering out identity-mismatched or weakly related groups.
    \item \textit{Technical Quality Assessment.} We remove samples with low Top-IQ scores \cite{chen2024topiq}. In addition, each image is temporarily encoded into JPEG format to estimate the bytes-per-pixel ratio, and images with abnormally low ratios are discarded because they often exhibit heavy compression artifacts or insufficient visual details.
\end{itemize}

\noindent
\subsection{Prompt Construction}

For large-scale prompt generation, we employ Qwen3-VL \cite{qwen3} to transform each filtered sample's metadata and extracted visual-textual signals into training prompts. The model takes available side information, such as user tags, source descriptions, OCR text, layout cues, and style labels, and produces descriptive prompts that better match the instruction format expected by the generation model. We also drop a small fraction of samples when automated prompting fails, or when the resulting text matches a preset list of sensitive terms.

The prompt construction process is designed to emphasize factuality, visual specificity, and controllable diversity across different task formats. For natural images, the generated prompts describe salient objects, attributes, spatial relations, scene context, and style. For graphic-layout and long-text samples, prompts additionally preserve the key textual content, reading order, and layout structure. For editing samples, Qwen3-VL is instructed to compare the source and target images and produce concise editing instructions that explain the intended visual transformation while avoiding unnecessary changes to preserved regions. For IP-oriented samples, it summarizes the identity-defining attributes of the reference person or object and constructs prompts that place the subject into new scenes while explicitly preserving its appearance. For multi-panel samples, Qwen3-VL describes both the global arrangement and the panel-level differences, enabling the model to learn grid composition and temporal-frame consistency. We also vary prompt granularity across short, medium, and detailed descriptions, so the final training corpus better reflects the diversity of real user inputs.

\begin{figure*}[!tb]
    \vspace{-0.0in}
    \centering {\includegraphics[width=1\textwidth]{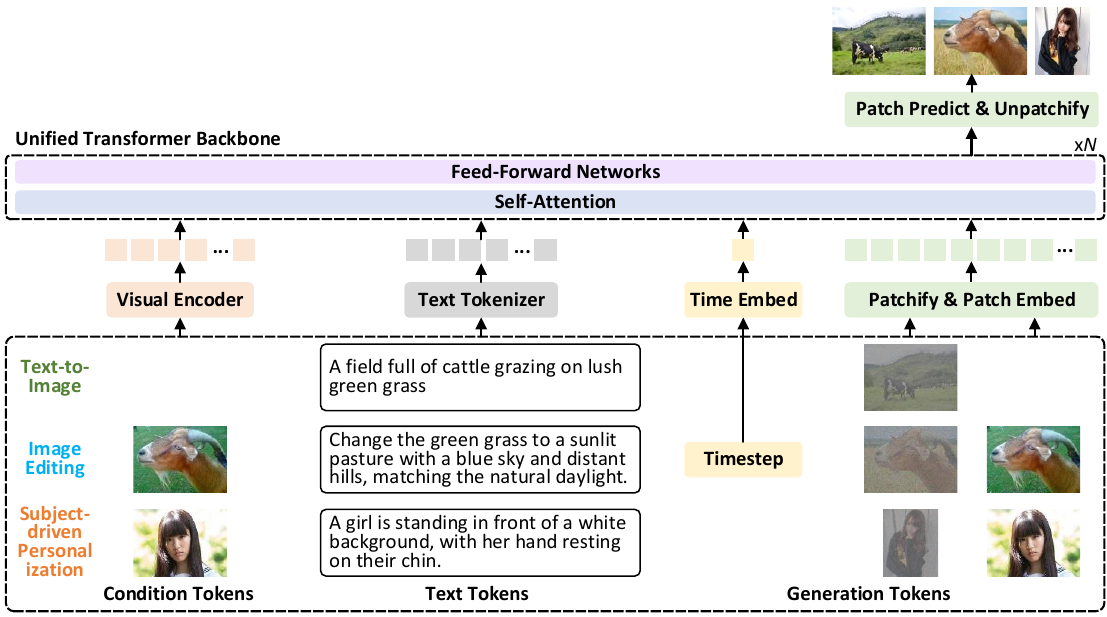}}
    \vspace{-0.0in}
    \caption{Overview of HiDream-O1-Image that enables a structural unification of multimodal inputs by mapping task-specific conditions, the text prompts, and raw pixels into a shared token space. These corresponding heterogeneous tokens (i.e., condition tokens, text tokens, and generation tokens formulated as noisy target samples along with timestep embeddings) are fed into the Unified Transformer backbone. In this way, HiDream-O1-Image treats diverse tasks (text-to-image, image editing, and subject-driven personalization) as an in-context reasoning process in the shared token space. Finally, the Transformer backbone predicts clean image patches which are reassembled to produce the target images.}
    \label{fig:framework}
    \vspace{-0.0in}
\end{figure*}

\section{Model Architecture: HiDream-O1-Image} \label{sec:model_architecture}
In this section, we introduce HiDream-O1-Image, a Pixel-level Unified Transformer that bridges the gap between high-fidelity pixel-space synthesis and versatile in-context reasoning for generalist image generation. Note that in order to demonstrate the structural scalability and versatility of our unified Pixel Diffusion Transformer paradigm, we instantiate HiDream-O1-Image at two distinct scales: an efficient 8B-parameter version for agile deployment and a massive 200B+ parameter version to push the boundaries of generation quality. 

Specifically, to effectively translate complex user intentions into model-preferred inputs, we first introduce our Reasoning-Driven Prompt Agent in Section \ref{tech-agent}. We then present an overview of our unified multimodal tokenization in Section \ref{tech-1}. Next, we detail the Unified Transformer (UiT) architecture of our HiDream-O1-Image, including both the backbone design and the hybrid Unified Attention mechanism, in Section \ref{tech-2}. Finally, the overall objective is introduced in Section \ref{tech-3}.

\subsection{Reasoning-Driven Prompt Agent}\label{tech-agent}
A significant bottleneck in current image generation models is the semantic gap between raw, often ambiguous user instructions and the dense, descriptive prompts required by the generation pipeline. To address this, we introduce a Reasoning-Driven Prompt Agent equipped with ``thinking'' mechanism built upon Gemma \cite{gemma4_2026}. When presented with a complex user query, the agent does not merely forward the text; instead, it explicitly reasons through the spatial layout, subject attributes, physical logic, and contextual relationships implied by the task. By explicitly generating a chain of thought before outputting the final prompt, the agent effectively refines and enriches the raw input. This reasoning process ensures that the subsequent HiDream-O1-Image receives highly unambiguous and structurally aligned textual conditions, significantly elevating the model's capability to handle intricate and reasoning-heavy visual generation and editing tasks.

\subsection{Unified Multimodal Tokenization}\label{tech-1}
The core of HiDream-O1-Image is a structural unification of heterogeneous modalities into a single shared token space. As illustrated in Figure~\ref{fig:framework}, during training, we define a comprehensive unified multimodal tokenization scheme to encode various inputs (i.e., the refined input text prompt \(y\), task-specific conditions \(c\), and target image \(x\)) into the shared token space. Specifically, we decompose the input stream into three primitive token types:
\begin{itemize}
\item \textbf{Text Tokens (\(y\)):} The refined text prompt \(y\), outputted by our Reasoning-Driven Prompt Agent, is converted into discrete tokens via the backbone's native vocabulary \cite{qwen3}, which are further mapped into the shared space.
\item \textbf{Condition Tokens (\(c\)):} For generation tasks requiring visual grounding (e.g., editing or subject-driven personalization), the input context images \(c\) (e.g., editing sources or reference subjects) are projected into semantic-rich tokens using a visual encoder (SigLip-2 \cite{siglip}). We further align the semantic-rich tokens to the shared space via a learnable projection.
\item \textbf{Generation Token (\(x_t\)):} For each target image \(x\), we construct the generation token \(x_t\) (i.e., noisy sample) via linear interpolation between the clean image \(x\) and Gaussian noise \(\epsilon \sim \mathcal{N}(0, I)\): \(x_t = t x + (1 - t)\epsilon\), where \(t\) denotes the diffusion timestep. The generation token is then partitioned into non-overlapping patches, and further projected into the shared space through a learnable patch embedding layer.
\end{itemize}
By concatenating all three kinds of tokens, we contextually encode them by a stack of unified Transformer blocks subsequently, which enables joint contextual reasoning across all modalities natively. Finally, a linear prediction head maps each output token back to the corresponding clean image patch, producing the reconstructed image estimations.

\subsection{Unified Transformer (UiT) Architecture}\label{tech-2}
\textbf{Backbone.} 
Our backbone is built upon a decoder-only Transformer architecture (i.e., a stack of unified Transformer blocks) inherited from large language models. To accommodate diverse application scenarios, we design two variants of HiDream-O1-Image: an 8B-parameter model and a 200B+ parameter model. Specifically, the 8B variant is initialized from a multimodel understanding backbone (Qwen3-VL-8B-Instruct \cite{qwen3}) to leverage its robust multimodal pre-alignment capability with high efficiency. Meanwhile, the 200B+ variant pushes the limits of the pixel-level Unified Transformer architecture by scaling to over 200 billion parameters, unlocking stronger capacity for complex visual reasoning and high-resolution synthesis. 

Both variants adopt RMSNorm \cite{rmsnorm} for normalization, SwiGLU \cite{swiglu} as the activation function, and RoPE \cite{rope} for positional encoding. To better inherit the pretrained autoregressive capability, we encode the diffusion timestep as an additional specialized token. To support the pixel-space diffusion process, we further incorporate learnable input and output patch embeddings into the backbone, as illustrated in Figure~\ref{fig:framework}. This approach enables direct modeling in pixel space without modifying the core Transformer structure across different scales.

\noindent\textbf{Hybrid Unified Attention Mechanism.} 
The causal attention paradigm is central to autoregressive language modeling, ensuring that each token attends only to preceding tokens in order to maintain the autoregressive property. In contrast, Diffusion Transformers for image synthesis typically adopt a full self-attention paradigm, allowing each visual token to attend to all other tokens to capture global spatial dependencies. 

In HiDream-O1-Image, we reconcile these two paradigms through a hybrid unified attention mechanism tailored for heterogeneous modalities. Concretely, the condition and text tokens follow causal masking and attend only to preceding multimodal tokens in the sequence. Generation tokens, by contrast, adopt full attention and can attend to all tokens, enabling global context aggregation during the diffusion process. Such a design elegantly preserves the autoregressive structure for language modeling while facilitating spatially coherent image synthesis within a unified Transformer framework.

\subsection{Overall Objective}\label{tech-3}
To achieve high-fidelity image synthesis in the raw pixel space, we adopt a joint optimization objective that balances structural regression with perceptual alignment. While the diffusion process in pixel space captures fine-grained spatial details, it often struggles to model long-range semantic coherence. Our strategy addresses this by coupling a flow matching loss for image prediction with perceptual supervision constraints (LPIPS \cite{lpips} loss and perceptual DINO loss).

\section{Model Training}\label{sec:training}
\subsection{Progressive Generalist Pre-training}
Here we scale HiDream-O1-Image through a three-stage progressive training strategy that transitions from foundational alignment to high-resolution generalist synthesis. Note that the training data are curated from coarse to fine, and the image resolution is gradually increased. Throughout all stages, we preserve the original aspect ratio of images to support flexible multi-resolution generation.

\noindent\textbf{Stage I: Foundational Alignment (512 \(\times\) 512).} 
In the first stage, we jointly optimize HiDream-O1-Image on three tasks: text-to-image generation (T2I), language modeling (LM), and multimodal understanding (MMU). This joint optimization is conducted over a mixture of image-text pairs and text-only corpora. As such, the model not only learns to semantically associate native pixel patches with linguistic concepts, but also retains strong linguistic capability. It is worthy to note that we adopt a relatively low image resolution (512 \(\times\) 512) and a large batch size in this stage, allowing the model to scale to billions of image-text pairs.

\noindent\textbf{Stage II: Generalist In-Context Learning (1,024 \(\times\) 1,024).} 
In the second stage, we enlarge the image resolution to 1,024 \(\times\) 1,024, aiming to enhance spatial fidelity and fine-grained detail generation in image synthesis. More importantly, we expand the training tasks (T2I, LM, and MMU) to include in-context generation and editing tasks (e.g., image editing and subject-driven personalization). This stage seamlessly integrates the Prompt Agent's explicit reasoning process with diverse synthesis scenarios, thereby strengthening reasoning-driven conditional generation and unified in-context learning.

\noindent\textbf{Stage III: High-Fidelity Refinement (2,048 \(\times\) 2,048).}
In this stage, the training of HiDream-O1-Image is restricted to an ultra-high-resolution subset with image resolutions exceeding 2,048 \(\times\) 2,048. This stage focuses exclusively on the refinement of fine-grained details and perceptual quality at ultra-high resolutions.

\subsection{Post-training}
Next, we conduct post-training optimization of our model in a two-stage paradigm, i.e., Supervised Fine-Tuning (SFT) and Reinforcement Learning from Human Feedback (RLHF), which together progressively refine both the reasoning capability and the generative quality.

\noindent\textbf{Stage I: SFT.} This stage aims to enhance visual aesthetics, photorealism, and prompt reasoning through a data-centric optimization strategy. Specifically, we construct a hybrid training corpus comprising several hundred thousand samples, which reflect high-quality compositional coherence, lighting consistency, photographic realism, and even stylistic fidelity across diverse artistic domains. Crucially, we also include high-quality reasoning trajectories to fine-tune the Prompt Agent, ensuring it consistently generates structurally aligned and unambiguous prompts. Moreover, we replace the Logit-Normal sampling strategy adopted in pre-training with uniform sampling, which ensures balanced timestep coverage and increases the effective training emphasis on late-stage denoising steps that capture fine-grained visual details.

\noindent\textbf{Stage II: RLHF.}
In this stage, we adopt GRPO \cite{liu2026flow} to further align the model with human preferences via reinforcement learning. In particular, we construct a composite advantage function by aggregating multiple reward signals produced by our reward models, including OCR accuracy, aesthetic assessment, instruction-following fidelity, and reasoning quality. This aggregated objective enables targeted improvements in photorealism, aesthetic quality, text rendering accuracy, semantic consistency, and logical reasoning, while effectively suppressing~artifacts.
   
   \begin{figure}
    \centering
    \includegraphics[width=0.7\linewidth]{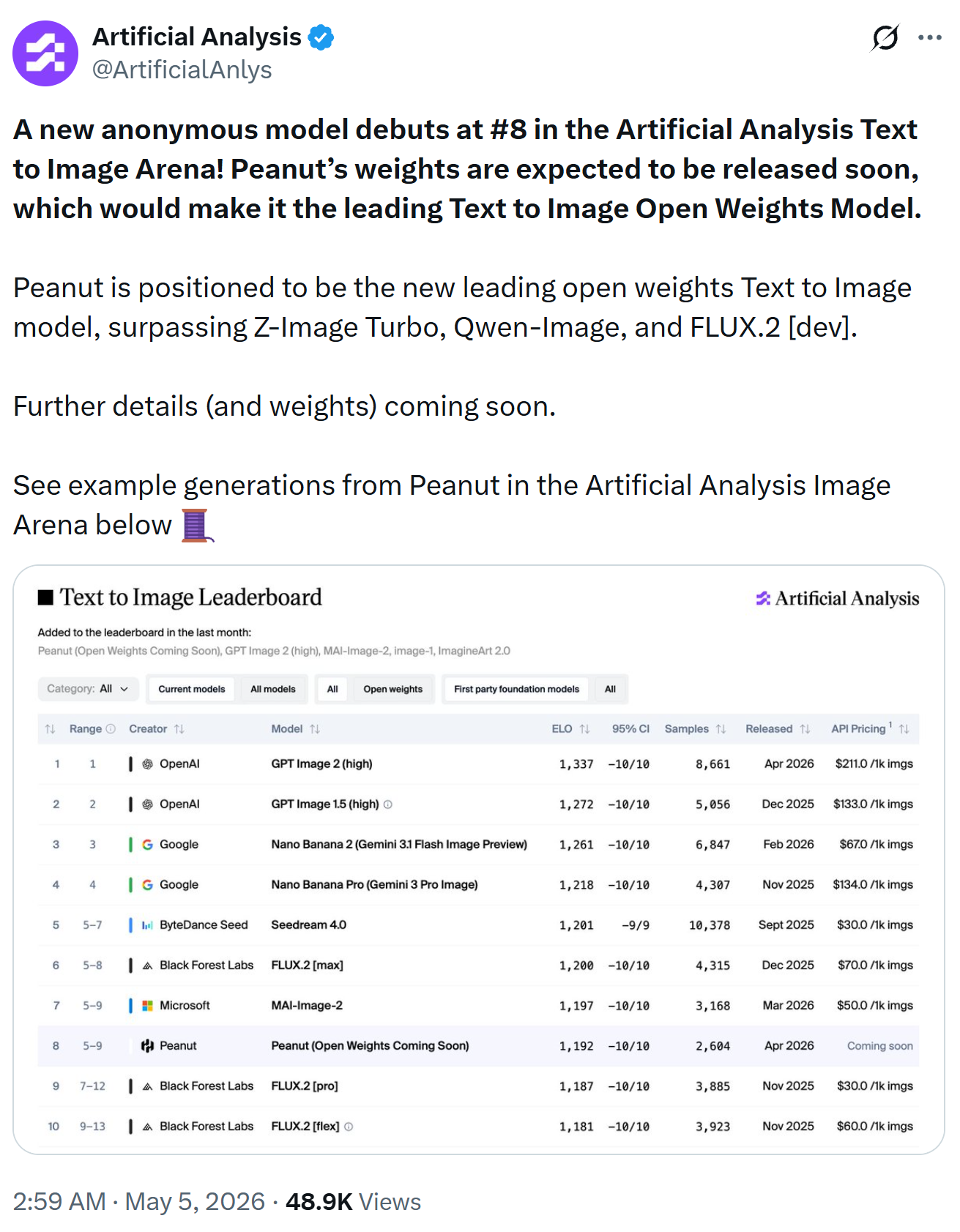} 
    \caption{HiDream-O1-Image (codename: Peanut) debuts at \#8 in the \href{https://artificialanalysis.ai/image/leaderboard/text-to-image}{Artificial Analysis Text to Image Arena}, which is positioned to be the new leading open weights text-to-image model (Date: 2026/5/5).}
    \label{fig:leaderboard}
  \end{figure}

\begin{table*}[t]
\setlength\tabcolsep{1.5pt}
\centering
\caption{Quantitative results on GenEval. Best results are highlighted in \textbf{bold}, and second-best results are \underline{underlined}.}
\begin{tabular}{l | c | c c c  c c c | c}
\toprule
\textbf{Model} & \textbf{\#Params} & \textbf{Single-Obj} & \textbf{Two-Obj} & \textbf{Count} & \textbf{Color} & \textbf{Position} & \textbf{Attr} & \textbf{Overall} \\
\midrule
Nano Banana 2.0~\cite{banana2024} & - & 1.00 & 0.96 & 0.71 & 0.84 & 0.86 & 0.65 & 0.83 
    \\ 
Seedream-4.0~\cite{seedream2025seedream} & - & 1.00 & 0.92 & 0.71 & 0.93 & 0.78 & 0.68 & 0.84 
  \\
GPT Image 1 [High]~\cite{gptimage2025} &   -    & 0.99  & 0.92  & 0.85  & 0.92  & 0.75  & 0.61  &  0.84  \\
GPT Image 2~\cite{gptimage2025} &   -    & 0.99 & 0.98 & 0.85 & 0.93 & 0.85 & 0.77 & 0.89  \\
\midrule
PixArt~\cite{chen2023pixart}  &   4.3B + 0.6B     & 0.98  & 0.50  & 0.44  & 0.80  & 0.08  & 0.07  &  0.48  \\
Show-o~\cite{xie2024show}   &  1.3B     & 0.95  & 0.52  & 0.49  & 0.82  & 0.11  & 0.28  &  0.53  \\
Emu3-Gen~\cite{wang2024emu3} &   8B    & 0.98  & 0.71  & 0.34  & 0.81  & 0.17  & 0.21  &  0.54  \\
SD3-Medium~\cite{sd3medium}   &  5.5B + 2B & 0.98  & 0.74  & 0.63  & 0.67  & 0.34  & 0.36  &  0.62  \\
JanusFlow~\cite{ma2025janusflow} &  1.3B    & 0.97  & 0.59  & 0.45  & 0.83  & 0.53  & 0.42  &  0.63  \\
FLUX.1 [Dev]~\cite{flux1}    &  4.8B + 12B    & 0.98  & 0.81  & 0.74  & 0.79  & 0.22  & 0.45  &  0.66  \\
SD3.5 Large ~\cite{sd3medium} &    5.5B + 8.1B      & 0.98  & 0.89  & 0.73  & 0.83  & 0.34  & 0.47  &  0.71  \\
Janus-Pro-7B~\cite{chen2025januspro7b} &  7B  & 0.99  & 0.89  & 0.59  & 0.90  & 0.79  & 0.66  &  0.80  \\
Z-Image-Turbo~\cite{team2025zimage} & 4B + 6B & 1.00 & 0.95 & 0.77 & 0.89 & 0.65 & 0.68 & 0.82   \\
FLUX.2 [Dev]~\cite{flux-2-2025}    & 24B + 32B & 1.00 & 0.99 & 0.79 & 0.93 & 0.73 & 0.78 & 0.87 \\
Qwen-Image~\cite{qwenimage} &  7B + 20B   & 0.99  & 0.92  & 0.89  & 0.88  & 0.76  & 0.77  &  0.87  \\ \midrule
\textbf{HiDream-O1-Image}  & 8B         & 1.00  & 0.99  & 0.79  & 0.89  & 0.93  & 0.78  &  \underline{0.90}  \\
\textbf{HiDream-O1-Image-Pro}  & 200B+ & 1.00 & 0.99 & 0.85 & 0.94 & 0.94 & 0.79 & \textbf{0.92} 
    \\
\bottomrule
\end{tabular}
\label{tab:GenEval}
\end{table*}

\begin{table}[t]
\setlength\tabcolsep{3.5pt}
\centering
\caption{Quantitative results on DPG. Best results are highlighted in \textbf{bold}, and second-best results are \underline{underlined}.}
\resizebox{\linewidth}{!}{%
\begin{tabular}{l|c|ccccc|c}
\toprule
\textbf{Model}    & \textbf{\#Params}     & \textbf{Global} & \textbf{Entity} & \textbf{Attribute} & \textbf{Relation} & \textbf{Other} & \textbf{Overall} \\
\midrule
GPT Image 1 {[}High{]}~\cite{gptimage2025} & - & 88.89 & 88.94 & 89.84 & 92.63 & 90.96 & 85.15 \\
GPT Image 2~\cite{gptimage2025} & - & 87.27 & 91.91 & 90.85 & 91.59 & 91.58 & 85.98 \\
Nano Banana 2.0~\cite{banana2024} & - & 85.17 & 92.55 & 91.16 & 90.45 & 91.08 & 86.90 \\ 
Seedream-4.0~\cite{seedream2025seedream} & - & 87.17 & 92.41 & 92.29 & 93.33 & 95.48 & 88.63 \\
\midrule
SD v1.5~\cite{rombach2022ldm}         &   0.12B + 0.86B    & 74.63           & 74.23           & 75.39              & 73.49             & 67.81          & 63.18    \\
PixArt~\cite{chen2023pixart}       &   4.3B + 0.6B     & 74.97           & 79.32           & 78.60              & 82.57             & 76.96          & 71.11      \\
Lumina-Next~\cite{zhuo2024lumina}  &  2B + 2B        & 82.82           & 88.65           & 86.44              & 80.53             & 81.82          & 74.63     \\
SDXL~\cite{podell2023sdxl}         &    0.81B + 2.6B      & 83.27           & 82.43           & 80.91              & 86.76             & 80.41          & 74.65    \\
Hunyuan-DiT~\cite{li2024hunyuandit}    &  4.8B + 1.5B      & 84.59           & 80.59           & 88.01              & 74.36             & 86.41          & 78.87    \\
Emu3-Gen~\cite{wang2024emu3}      &    8B     & 85.21           & 86.68           & 86.84              & 90.22             & 83.15          & 80.60    \\
DALL-E 3~\cite{openai2023dalle3}      &    -     & 90.97           & 89.61           & 88.39              & 90.58             & 89.83          & 83.50     \\
FLUX.1{[}Dev{]}~\cite{flux1}  &  4.8B + 12B    & 74.35           & 90.00           & 88.96              & 90.87             & 88.33          & 83.84     \\
SD3 Medium~\cite{sd3medium}     &   5.5B + 2B     & 87.90           & 91.01           & 88.83              & 80.70             & 88.68          & 84.08     \\
Janus-Pro-7B~\cite{chen2025januspro7b}   &   7B     & 86.90           & 88.90           & 89.40              & 89.32             & 89.48          & 84.19     \\
Z-Image-Turbo~\cite{team2025zimage} & 4B + 6B & 91.29 & 89.59 & 90.14 & 92.16 & 88.68 & 84.86  \\
HiDream-I1-Full~\cite{cai2025hidream}  &  13.5B + 17B    & 76.44           & 90.22           & 89.48              & 93.74             & 91.83          & 85.89     \\
FLUX.2 [Dev]~\cite{flux-2-2025}    & 24B + 32B & 92.20 & 91.36 & 93.28 & 93.52 & 89.72 & 87.57 
 \\
Qwen-Image~\cite{qwenimage}    &    7B + 20B     & 91.32           & 91.56           & 92.02              & 94.31             & 92.73          & 88.32      \\
\midrule
\textbf{HiDream-O1-Image}  &   8B      & 95.15           & 92.32           & 93.74              & 92.88             & 90.25          & \underline{89.83}      \\ 
\textbf{HiDream-O1-Image-Pro}  & 200B+ & 94.97 & 95.42 & 92.59 & 90.82 & 89.50 & \textbf{90.30} 
      \\
\bottomrule
\end{tabular}
}
\label{tab:dpg}
\vspace{-0.1in}
\end{table}

\begin{table*}[!tb] \scriptsize
\setlength\tabcolsep{1.3pt}
\centering
\caption{Quantitative results on HPSv3. Best results are highlighted in \textbf{bold}, and second-best results are \underline{underlined}.}
\begin{tabular}{l|c|c|cccccccccccc}
\toprule
\textbf{Models}            & \textbf{\#Params}   & \textbf{All}   & \begin{tabular}[c]{@{}c@{}}\textbf{Chara}\\ \textbf{cters}\end{tabular} & \textbf{Arts}  & \textbf{Design} & \begin{tabular}[c]{@{}c@{}}\textbf{Archite}\\ \textbf{chture}\end{tabular} & \textbf{Animals} & \begin{tabular}[c]{@{}c@{}}\textbf{Natural}\\ \textbf{Scenery}\end{tabular} & \begin{tabular}[c]{@{}c@{}}\textbf{Transpo}\\ \textbf{rtation}\end{tabular} & \begin{tabular}[c]{@{}c@{}}\textbf{Prod}\\ \textbf{ucts}\end{tabular} & \textbf{Plants} & \textbf{Food}  & \textbf{Science} & \textbf{Others} \\ \hline
Seedream-4.0~\cite{seedream2025seedream} & - & 9.32 & 9.83 & 9.20 & 8.83 & 9.95 & 8.99 & 9.40 & 9.58 &  	9.12 & 9.26 & 9.75 & 9.11 & 9.51 \\
Nano Banana 2.0~\cite{banana2024}   & -         & 10.01 & 10.18                                                 & 9.18  & 9.58   & 10.96                                                    & 9.71    & 10.04                                                     & 10.38                                                     & 10.36                                               & 10.14  & 10.61 & 9.14    & 9.89   \\
GPT Image 2~\cite{gptimage2025} & - & 10.21 & 10.75 & 9.91 & 10.15 & 10.59 & 10.05 & 10.29 & 10.17 &  	10.26 & 10.07 & 10.75 & 10.05 & 10.00 
 \\\midrule
Z-Image-Turbo~\cite{team2025zimage}     & 4B + 6B   & 8.35  & 8.98                                                  & 8.29  & 7.65   & 9.26                                                     & 8.51    & 8.33                                                      & 8.81                                                      & 7.83                                                & 8.46   & 8.64  & 7.93    & 8.57   \\
FLUX.2 [Dev]~\cite{flux-2-2025}      & 24B + 32B & 9.28  & 10.23                                                 & 9.56  & 8.80   & 9.73                                                     & 9.43    & 9.21                                                      & 9.44                                                      & 8.93                                                & 9.23   & 9.82  & 8.67    & 9.11   \\
Qwen-image~\cite{qwenimage}        & 7B + 20B  & 9.94  & 10.91                                                 & 10.47 & 9.56   & 10.22                                                    & 10.61   & 9.87                                                      & 10.10                                                     & 9.15                                                & 9.99   & 10.08 & 9.19    & 9.83   \\\hline
\textbf{HiDream-O1-Image}   & 8B        & \underline{10.37} & 10.59                                                 & 10.44 & 10.29  & 11.02                                                    & 10.34   & 10.37                                                     & 10.54                                                     & 10.50                                               & 10.38  & 10.85 & 9.68    & 10.09  \\
\textbf{HiDream-O1-Image-Pro} & 200B+     & \textbf{10.47} & 10.63                                                 & 10.51 & 10.33  & 11.11                                                    & 10.08   & 10.45                                                     & 10.37                                                     & 10.75                                               & 10.29  & 11.13 & 10.09   & 10.39  \\ \bottomrule
\end{tabular}
\label{tab:hpsv3}
\end{table*}

\section{Adversarial Diffusion Distillation for Fast Inference}
\label{sec:acceleration}

The full version of HiDream-O1-Image typically adopts around 50 denoising steps during inference. While this sampling configuration ensures strong visual quality, the iterative process may limit practical deployment in latency-sensitive scenarios. To improve inference efficiency, we further distill the full model into an accelerated variant with a shorter sampling trajectory \cite{yin2024improved}. Specifically, we construct \textbf{HiDream-O1-Image-Dev} as the efficient variant of HiDream-O1-Image, which adopts a 28-step sampler for faster generation.

The distillation process trains the student model to approximate the generation behavior of the teacher model under a reduced number of steps. We adopt DMD \cite{yin2024improved} as the core objective, denoted as $\mathcal{L}_{\text{DMD}}$, to align the trajectory distribution predicted by the student with that of the full HiDream-O1-Image model. This objective allows HiDream-O1-Image-Dev to inherit the main generative dynamics of the teacher while using a much shorter sampling schedule. In addition, we keep the standard diffusion loss as an auxiliary supervision term, which improves training stability and mitigates optimization oscillation during distillation.

To further preserve perceptual fidelity and image sharpness in HiDream-O1-Image-Dev, we incorporate adversarial learning into the distillation framework. The student model is regarded as the generator and is optimized together with a discriminator network. The discriminator compares real images with the images reconstructed from the pixel-space predictions of the student, and its classification is guided by multi-level features extracted from the frozen teacher backbone. The final objective of this GAN-powered distillation is formulated as a weighted combination of DMD, the standard diffusion loss, and the adversarial loss:
$\mathcal{L}_{\text{total}} = \mathcal{L}_{\text{DMD}} + \lambda_{\text{diff}} \mathcal{L}_{\text{diff}} + \lambda_{\text{adv}} \mathcal{L}_{\text{adv}}$.


\begin{table}[t]
\setlength\tabcolsep{3.5pt}
\centering
\caption{Quantitative results on CVTG-2K. Best results are highlighted in \textbf{bold}, and second-best results are \underline{underlined}.}
\resizebox{\linewidth}{!}{%
\begin{tabular}{l|c|ccccc|c|c}
\toprule
\multirow{2}{*}{\textbf{Model}} & \multirow{2}{*}{\textbf{\parbox{1cm}{\centering\#\\Params}}} & \multicolumn{5}{c|}{\textbf{Word Accuracy}}             & \multirow{2}{*}{\textbf{NED}} & \multirow{2}{*}{\textbf{\shortstack{CLIP\\Score}}} \\ \cmidrule{3-7}
                      &          & \parbox{1cm}{\centering2\\regions} & \parbox{1cm}{\centering3\\regions} & \parbox{1cm}{\centering4\\regions} & \parbox{1cm}{\centering5\\ regions} & average &                               &                                     \\ \midrule
Nano Banana 2.0~\cite{banana2024} & - & 0.7465 & 0.7720 & 0.8067 & 0.7980 & 0.7875 & 0.8945 & 0.7212     \\ 
GPT Image 1 {[}High{]}~\cite{gptimage2025}  &  -      & 0.8779    & 0.8659    & 0.8731    & 0.8218    & 0.8569  & 0.9478 & 0.7982 \\
Seedream-4.0~\cite{seedream2025seedream} & - & 0.8980 & 0.8949 & 0.9044 & 0.9015 & 0.9003 & 	0.9511 & 0.8033 
  \\
GPT Image 2~\cite{gptimage2025}  &  -      & 0.8904 & 0.8887 & 0.9101 & 0.9044 & 0.9003 & 0.9515 & 0.7798                               \\
\midrule
TextDiffuser-2~\cite{chen2024textdiffuser2}    &   0.12B + 0.9B           & 0.5322    & 0.3255    & 0.1787    & 0.0809    & 0.2326  & 0.4353                        & 0.6765                              \\
RAG-Diffusion~\cite{chen2024region}     &   4.8B + 12B  & 0.4388    & 0.3316    & 0.2116    & 0.1910    & 0.2648  & 0.4498                        & 0.7797                              \\
AnyText~\cite{tuo2024anytext}           &  0.123B + 1.2B            & 0.0513    & 0.1739    & 0.1948    & 0.2249    & 0.1804  & 0.4675                        & 0.7432                              \\
3DIS~\cite{zhou20243dis}             &   0.81B + 2.6B              & 0.4495    & 0.3959    & 0.3880    & 0.3303    & 0.3813  & 0.6505                        & 0.7767                              \\
FLUX.1 {[}dev{]}~\cite{flux1}     &   4.8B + 12B        & 0.6089    & 0.5531    & 0.4661    & 0.4316    & 0.4965  & 0.6879                        & 0.7401                              \\
SD3.5 Large~\cite{sd3medium}           &   5.5B + 8.1B       & 0.7293    & 0.6825    & 0.6574    & 0.5940    & 0.6548  & 0.8470                        & 0.7797                              \\
TextCrafter~\cite{du2025textcrafter}       &    7B + 20B          & 0.7628    & 0.7628    & 0.7406    & 0.6977    & 0.7370  & 0.8679                        & 0.7868                              \\
Qwen-Image~\cite{qwenimage}            &  7B + 20B        & 0.8370    & 0.8364    & 0.8313    & 0.8158    & 0.8288  & 0.9116                        & 0.8017                             \\
Z-Image-Turbo~\cite{team2025zimage} & 4B + 6B & 0.8872 & 0.8662 & 0.8628 & 0.8347 & 0.8585 & 0.9281 & 0.8048   \\
FLUX.2 [Dev]~\cite{flux-2-2025}    & 24B + 32B & 0.9261 & 0.8897 & 0.8995 & 0.8732 &0.8926 & 0.9475 & \underline{0.8104} \\ \midrule
\textbf{HiDream-O1-Image}    & 8B                & 0.9085    & 0.9159    & 0.9216    & 0.9015   & \underline{0.9128}  & \underline{0.9561}                        & {0.8076}                              \\
\textbf{HiDream-O1-Image-Pro}    & 200B+ & 0.9133 & {0.9221} & 0.9365 & 0.9175 & \textbf{0.9222} & \textbf{0.9628} & \textbf{0.8349} 
\\
\bottomrule
\end{tabular}
\label{tab:cvtg}
}
\end{table}

\begin{table}[!tb]
\setlength\tabcolsep{3.5pt}
\centering
\caption{Quantitative results on LongText-Bench. Best results are highlighted in \textbf{bold}, and second-best results are \underline{underlined}.}
\begin{tabular}{l|c|c|c}
\toprule
\textbf{Model}         & \textbf{\#Params}         & \textbf{LongText-Bench-EN} & \textbf{LongText-Bench-ZH} \\ \midrule
Seedream-4.0~\cite{seedream2025seedream} & - & 0.936 & 0.946  \\
GPT Image 1 {[}High{]}~\cite{gptimage2025} &  - & 0.956             & 0.619             \\
GPT Image 2~\cite{gptimage2025} &  - & 0.960             & 0.961             \\
Nano Banana 2.0~\cite{banana2024} & - & \underline{0.980} & 0.965   \\ 
\midrule
Janus-Pro-7B~\cite{chen2025januspro7b}     &   7B      & 0.019             & 0.006             \\
BLIP3-o~\cite{chen2025blip3}        &  7B + 1.4B       & 0.021             & 0.018             \\
Kolors 2.0~\cite{Kolors2}    &    -      & 0.258             & 0.329             \\
BAGEL~\cite{bagle}       &   7B + 7B         & 0.373             & 0.310             \\
OmniGen2~\cite{wu2025omnigen2}    &  3B + 4B         & 0.561             & 0.059             \\
X-Omni~\cite{geng2025xomni}      &  7B           & 0.900             & 0.814             \\
HiDream-I1-Full~\cite{cai2025hidream} &  13.5B + 17B      & 0.543             & 0.024             \\
FLUX.1 {[}Dev{]}~\cite{flux1} &   4.8B + 12B    & 0.607             & 0.005             \\
Z-Image-Turbo~\cite{team2025zimage} & 4B + 6B & 0.917 & 0.926  \\
FLUX.2 [Dev]~\cite{flux-2-2025}    &  24B + 32B  & 0.963 &	0.757   \\
Qwen-Image~\cite{qwenimage}    &      7B + 20B    & 0.943             & 0.946             \\ \midrule
\textbf{HiDream-O1-Image}  & 8B         & {0.979}             & \underline{0.978}             \\
\textbf{HiDream-O1-Image-Pro}  & 200B+ & \textbf{0.982} & \textbf{0.980}\\
\bottomrule
\end{tabular}
\label{tab:longtext}
\end{table}

\section{Performance Comparisons for Text-to-Image Generation}
Here we systematically evaluate our HiDream-O1-Image's text-to-image generation capabilities, spanning from general visual synthesis to fine-grained text~rendering.

\subsection{General Text-to-Image Synthesis}
We first benchmark general T2I synthesis performance on GenEval~\cite{ghosh2023geneval}, DPG~\cite{hu2024dpg}, and HPSv3~\cite{ma2025hpsv3} datasets.
As quantitatively reported in Table~\ref{tab:GenEval}, Table~\ref{tab:dpg}, and Table~\ref{tab:hpsv3}, our HiDream-O1-Image model with 8B parameters significantly surpasses existing open-source counterparts of similar scale (e.g., Z-Image-Turbo, SD3.5 Large, Janus-Pro-7B). Furthermore, our scaled-up 200B+ model achieves state-of-the-art fidelity, outperforming leading closed-source models such as GPT Image 2 and Seedream-4.0.
The results generally highlight the key advantage of our structural unification design in HiDream-O1-Image. 
By projecting both vision and language into a natively shared token space, HiDream-O1-Image bypasses the semantic gap inherent in conventional paradigms that employ disjoint text encoders and image generators, thereby achieving superior cross-modality alignment.

\subsection{High-Fidelity Text Rendering}

We further rigorously examine the model's text rendering capabilities on CVTG-2K~\cite{du2025textcrafter} and LongText-Bench~\cite{geng2025xomni} datasets.
As detailed in Table~\ref{tab:cvtg} and Table~\ref{tab:longtext}, HiDream-O1-Image attains the highest scores across most metrics on CVTG-2K.
In addition, on LongText-Bench, our 8B model shows comparable performance with the best competitor Qwen-Image with heavier parameters (27B), while our 200B+ model further pushes the boundaries of ultra-long text rendering, establishing a new state-of-the-art.
This clear leap in character-level accuracy stems directly from our end-to-end pixel-space generative framework:
By circumventing the intermediate text-to-vision translation bottlenecks inherent in disjoint modality encoding and bypassing lossy VAE compression, our model achieves precise text-image alignment while mitigating structural distortions in visual text rendering.

\subsection{Versatility Across Diverse Generation Scenarios}
\begin{figure}
    \centering
    \includegraphics[width=1\linewidth]{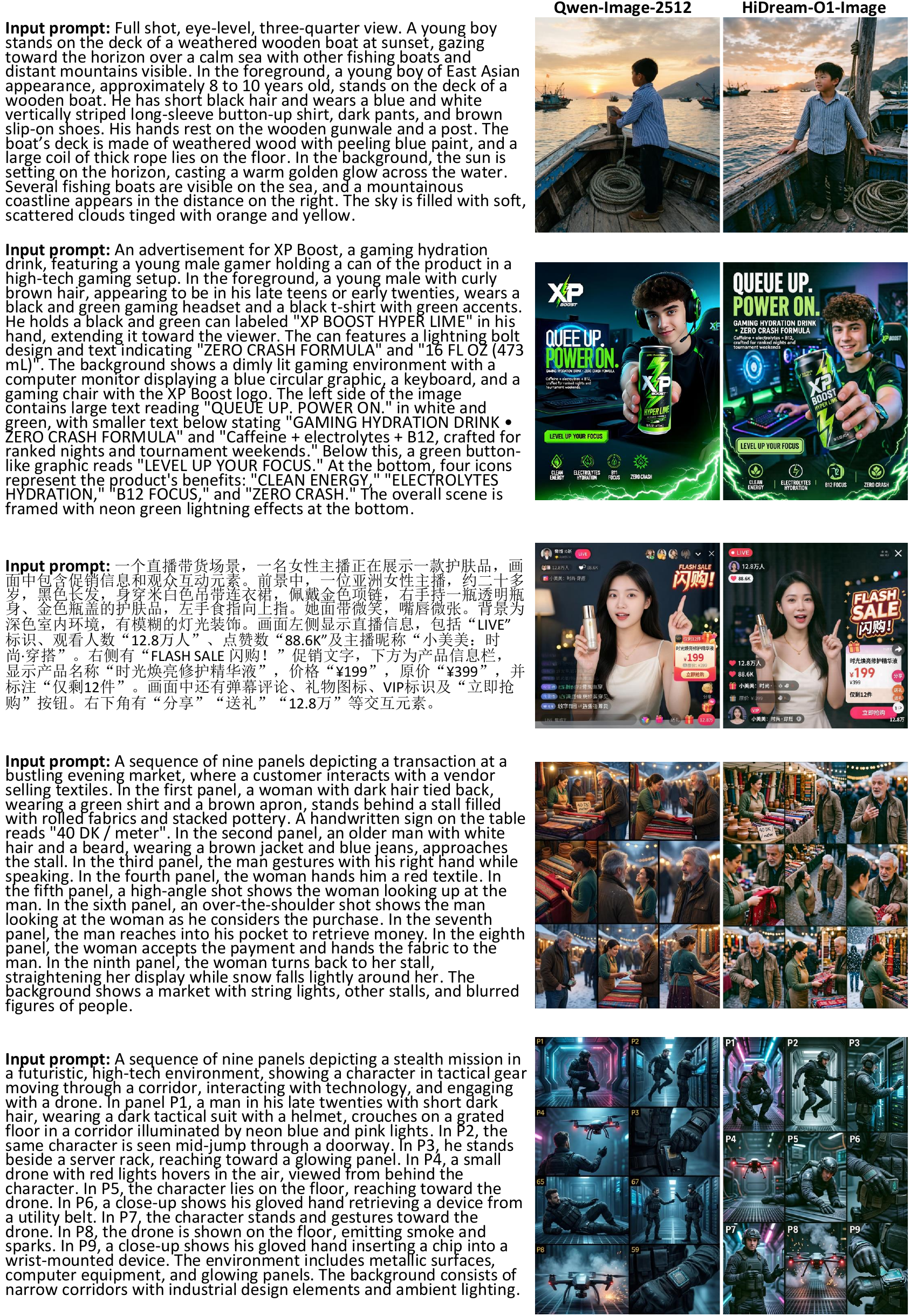} 
    \caption{Comparison between Qwen-Image and HiDream-O1-Image on text-to-image capability for diverse cinematic shots, complex text rendering, and multi-panel image  scenarios.}
    \label{fig:gen_images_qwen}
  \end{figure}

Beyond standard quantitative benchmarks, HiDream-O1-Image demonstrates strong versatility across various practical text-to-image generation scenarios. As visually illustrated in Figure~\ref{fig:gen_images_qwen}, HiDream-O1-Image seamlessly handles diverse cinematic shots, versatile artistic styles, complex long text rendering, and multi-panel image generation for storyboard~production. 

Crucially, we recognize that high-quality static image generation often serves as the foundational entry point (i.e., the initial keyframe) for downstream video generation. To this end, HiDream-O1-Image places a particular emphasis on mastering cinematic camera language and structural storyboard layouts. By ensuring rigorous controllability over spatial composition and camera perspectives at the initial image generation stage, our model provides a controllable visual anchor, which significantly benefits and streamlines subsequent video synthesis tasks.

Specifically, to facilitate professional cinematic control, HiDream-O1-Image natively supports fine-grained manipulation across 15 distinct cinematic shots and camera perspectives. These comprehensively encompass:
\begin{itemize}
    \item \textbf{Shot Scales:} extreme full shot, full shot, medium full shot, medium shot, medium close-up, close-up, and extreme close-up.
    \item \textbf{Camera Angles:} high angle, low angle, eye-level, and bird's-eye view.
    \item \textbf{Subject Orientations:} front view, side view, back view, and three-quarter view.
\end{itemize}

Furthermore, its robust capability in multi-panel image generation enables the coherent creation of storyboards within a single inference pass. This comprehensive scenario coverage ensures that HiDream-O1-Image is not merely a static image generator, but a versatile visual engine tailored for cinematic pre-production and video generation workflows.

\section{Performance Comparisons for Image Editing}
We comprehensively evaluate HiDream-O1-Image's editing performance on two diverse benchmarks (GEdit~\cite{liu2025step1xEdit} and ImgEdit~\cite{ye2025imgedit}). Quantitative results in Table~\ref{tab:gedit} and Table~\ref{tab:imgedit} demonstrate that our model achieves strong instruction-following capability while preserving high generation fidelity. Notably, with only 8B parameters, our HiDream-O1-Image yields comparable results against several significantly larger competitors (e.g., 16.8B FLUX.1 Kontext and 27B Qwen-Image-Edit). Furthermore, our scaled-up 200B+ model establishes a new state-of-the-art, outperforming top-tier proprietary models in handling highly complex, fine-grained manipulations. This basically validates the effectiveness of our core methodological choice: unifying visual and linguistic inputs within a natively shared token space intrinsically dissolves the modality barrier and encourages precise grounding of semantic concepts to their corresponding spatial regions, triggering an effective in-context generation process. Additionally, the joint language modeling and multimodal understanding objectives enforced during pre-training equip the model to accurately interpret user intents while preserving semantic awareness of the original visual context for detail fidelity. Together, these capabilities encourage our framework to capture nuanced editing intents, execute highly precise manipulations, and ensure faithful preservation of unedited regions across both model scales.

\begin{table}[t]
\centering
\caption{Quantitative results on GEdit. Best results are highlighted in \textbf{bold}, and second-best results are \underline{underlined}.}
\renewcommand{\arraystretch}{0.95}
\begin{tabular}{l |c| c c c}
\toprule
\textbf{Model} & \textbf{{\centering\#Params}} & \textbf{Q-SC} & \textbf{Q-PQ} & \textbf{Q-O} \\ 
\midrule
Nano Banana 2.0~\cite{banana2024} & - & 7.66 & 7.46 & 7.32 \\
Seedream-4.0~\cite{seedream2025seedream} & - & 7.84 & 7.50 & 7.53 \\
GPT Image 2~\cite{gptimage2025} &   - & 7.94 & 7.60 & 7.67 \\
\midrule
ICEdit~\cite{zhang2025icedit} &  4.8B + 12B & 5.10 & 7.05 & 4.96 \\
OmniGen~\cite{xiao2025omnigen} & 3.8B & 6.26 & 6.94 & 5.90 \\
FLUX.1 Kontext~\cite{labs2025flux1kontext} & 4.8B + 12B & 6.62 & 7.22 & 6.34 \\
OmniGen2~\cite{wu2025omnigen2} & 3B + 4B & 6.34 & 7.05 & 6.04 \\
BAGEL~\cite{bagle} & 7B + 7B & 7.29 & 6.57 & 6.64 \\
HiDream-E1.1~\cite{cai2025hidream} & 13.5B + 17B & 7.63 & 7.02 & 7.16 \\
Qwen-Image-Edit~\cite{qwenimage} & 7B + 20B & 7.76 & 7.47 & 7.41 \\ 
\midrule
\textbf{HiDream-O1-Image} & 8B & 7.99  & 7.42 & \underline{7.60} \\
\textbf{HiDream-O1-Image-Pro} & 200B+ & 8.05  & 7.47 & \textbf{7.67} \\
\bottomrule
\end{tabular}%

\label{tab:gedit}
\end{table}

\begin{table}[t]
\centering
\caption{Quantitative results on ImgEdit. Best results are highlighted in \textbf{bold}, and second-best results are \underline{underlined}.}
\footnotesize
\setlength{\tabcolsep}{1.5pt}
\renewcommand{\arraystretch}{0.95}
\begin{tabular}{l |c| c c c c c c c c c c}
\toprule
\textbf{Model} & \textbf{{\centering\#Params}} & \textbf{Add} & \textbf{Adjust} & \textbf{Extract} & \textbf{Replace} & \textbf{Remove} & \textbf{Background} & \textbf{Style} & \textbf{Hybrid} & \textbf{Action} & \textbf{Overall} \\
\midrule
Seedream-4.0~\cite{seedream2025seedream} & - & 4.67 & 4.55 & 2.65 & 4.78 & 4.52 & 4.53 & 4.78 & 3.51 & 4.66 & 4.29 \\
Nano Banana 2.0~\cite{banana2024} & - & 4.56 & 4.41 & 4.04 & 4.62 & 4.74 & 4.54 & 4.87 & 4.01 & 4.81 & \underline{4.51} \\
GPT Image 2~\cite{gptimage2025} &   - & 4.85 & 4.82 & 4.07 & 4.92 & 4.75 & 4.88 & 4.96 & 4.47 & 4.85 & \textbf{4.73}\\
\midrule
ICEdit~\cite{zhang2025icedit} &  4.8B + 12B & 3.58 & 3.39 & 1.73 & 3.15 & 2.93 & 3.08 & 3.84 & 2.04 & 3.68 & 3.05 \\
OmniGen~\cite{xiao2025omnigen} & 3.8B & 3.47 & 3.04 & 1.71 & 2.94 & 2.43 & 3.21 & 4.19 & 2.24 & 3.38 & 2.96 \\
FLUX.1 Kontext~\cite{labs2025flux1kontext} & 4.8B + 12B & 4.25 & 4.15 & 2.35 & 4.56 & 3.57 & 4.26 & 4.57 & 3.68 & 4.63 & 4.00 \\
OmniGen2~\cite{wu2025omnigen2} & 3B + 4B & 3.57 & 3.06 & 1.77 & 3.74 & 3.20 & 3.57 & 4.81 & 2.52 & 4.68 & 3.44 \\
BAGEL~\cite{bagle} & 7B + 7B & 3.56 & 3.31 & 1.70 & 3.30 & 2.62 & 3.24 & 4.49 & 2.38 & 4.17 & 3.20 \\
Qwen-Image-Edit~\cite{qwenimage} & 7B + 20B & 4.38 & 4.16 & 3.43 & 4.66 & 4.14 & 4.38 & 4.81 & 3.82 & 4.69 & {4.27} \\ 
\midrule
\textbf{HiDream-O1-Image} & 8B & 4.40 & 4.53 & 2.05 & 4.69 & 3.94 & 4.41 & 4.87 & 3.84 & 4.51 & 4.14 \\
\textbf{HiDream-O1-Image-Pro} & 200B+ & 4.48 & 4.57 & 4.35 & 4.74 & 4.66 & 4.49 & 4.83 & 3.8 & 	4.71 & \underline{4.51}
 \\
\bottomrule
\end{tabular}%

\label{tab:imgedit}
\end{table}

\begin{table}[t]
\caption{Quantitative results on UniSubject. Best results are highlighted in \textbf{bold}, and second-best results are \underline{underlined}.}
\scriptsize
\setlength\tabcolsep{1.5pt}
\centering
\begin{tabular}{l |c | c c c c| c c c c | c c c c} 
\toprule
& \multirow{2}{*}{\textbf{\parbox{1.2cm}{\centering\#\\Params}}} &\multicolumn{4}{c|}{\textbf{2-3 Subjects}} & \multicolumn{4}{c|}{\textbf{4-8 Subjects}} & \multicolumn{4}{c}{\textbf{9-11 Subjects}}\\ 
\textbf{Model} & & \textbf{Q-PF} & \textbf{Q-SC} & \textbf{Q-O} & \textbf{HPSv3} & \textbf{Q-PF} & \textbf{Q-SC} & \textbf{Q-O} & \textbf{HPSv3}  & \textbf{Q-PF} & \textbf{Q-SC} & \textbf{Q-O} & \textbf{HPSv3}  \\\midrule
\multicolumn{1}{l|}{Nano Banana 2.0~\cite{banana2024}}           &    -  &  8.81  & 8.19 & \textbf{8.50}  & \textbf{11.08}  & 8.35     & 7.09     & 7.72     & 9.51     & 8.63 & 7.33 & \textbf{7.98} & 9.26       \\
\multicolumn{1}{l|}{Seedream-4.0~\cite{seedream2025seedream}}   & - &   8.36 & 7.96 & {8.16} & 10.45 & 8.19 & 7.83 & \textbf{8.01} & \underline{9.53} & 8.23 & 7.49 & 7.86 & \underline{9.78}\\
\multicolumn{1}{l|}{GPT Image 2~\cite{gptimage2025}} &   - & 8.73 & 7.66 & \underline{8.20} & 11.91 & 8.66 & 6.39 & 7.52 & 10.89 & 8.97 & 5.56 & 7.26 & 11.33  \\
       \midrule
\multicolumn{1}{l|}{BAGEL~\cite{bagle}}  & 7B + 7B   &  7.85    &  4.35    & 6.10  & 8.63  & 8.18 & 4.16 & 6.17 & 7.79 & 8.16 & 4.01 & 6.09 & 7.99      \\
\multicolumn{1}{l|}{OmniGen2~\cite{wu2025omnigen2}}  & 3B + 4B &   7.93   &  6.39    & 7.16 &  10.41 &  7.97    & 5.32 & 6.65 & 9.21 & 8.03 & 4.47 & 6.25  & 9.32       \\
\multicolumn{1}{l|}{Qwen-Image-Edit~\cite{qwenimage}}  & 7B + 20B &  8.19    &   6.82   & 7.50 & 8.84  & 5.92 & 4.75 & 5.34 & 5.40   & 3.20  & 2.21 & 2.71 & 2.13  \\
\multicolumn{1}{l|}{DreamOmini2~\cite{xia2025dreamomni2}} & 4.8B + 19B &  7.48   &  6.31    &  6.90 & {10.10}  &     7.20   &  4.51    &  5.86   &   8.16   &   6.90   & 4.02 & 5.46  &  8.57    \\
\multicolumn{1}{l|}{Scone~\cite{wang2025scone}}   &  7B + 7B   &  7.65    &  6.65    &  7.15 & 8.97  &    7.22   & 6.01  &  6.62  &   7.74        &   6.61  & 4.94  & 5.78 & 7.54   \\
\multicolumn{1}{l|}{Echo-4o~\cite{ye2025echo}} &  7B + 7B &  8.08 & 6.84 &  {7.46} &  9.99   &  8.09    & 6.29    &   {7.19}   &   {8.61}   & 7.99 & 5.46  &  {6.73} &  {8.78}  \\
\midrule

\multicolumn{1}{l|}{\textbf{HiDream-O1-Image}}  & 8B  &    8.65  &  7.25  &  {7.95}   & {10.45} & 8.14  &  6.81   & \textbf{7.47}   & \underline{9.53}  &  8.35 & 6.95 & {7.65} &   \underline{9.78}  \\
\multicolumn{1}{l|}{\textbf{HiDream-O1-Image-Pro}}  & 200B+  &   8.87   &  8.12  &   \textbf{8.50}  & \underline{11.05} & 8.31 &   7.66  &  \underline{7.99}  & \textbf{9.76}  & 8.56 & 7.28 & \underline{7.92} & \textbf{9.83}    \\
    \bottomrule
\end{tabular}
\label{tab:unisubj}
\end{table}

\section{Performance Comparisons for Subject-driven Personalization}

Subject-driven customized generation aims to naturally recompose user-provided reference objects into novel contextual scenes. 
Here we curate \textbf{UniSubject}, a new test set designed to evaluate the model's capability in preserving and composing multiple subjects. UniSubject comprises 300 test cases, encompassing a total of 1.8K subjects. Each case pairs 1 human subject with 1 to 10 reference objects (e.g., clothes, car, furniture), accompanied by a compositional prompt. Following the practices of VIEScore~\cite{ku2024viescore}, we employ Qwen-VL2.5-72B to assess three metrics: \textbf{Prompt Following (Q-PF)} (alignment between the prompt and the generated image), \textbf{Subject Consistency (Q-SC)} (subject preservation between each reference image and the generated image in a pairwise manner), and the \textbf{Overall Score (O)} (the mean of Q-PF and Q-SC). Moreover, \textbf{HPSv3}~\cite{ma2025hpsv3} is adopted to measure human preference.

\begin{figure}[t]
    \vspace{-0.2in}
    \centering
    \includegraphics[width=1.0\linewidth]{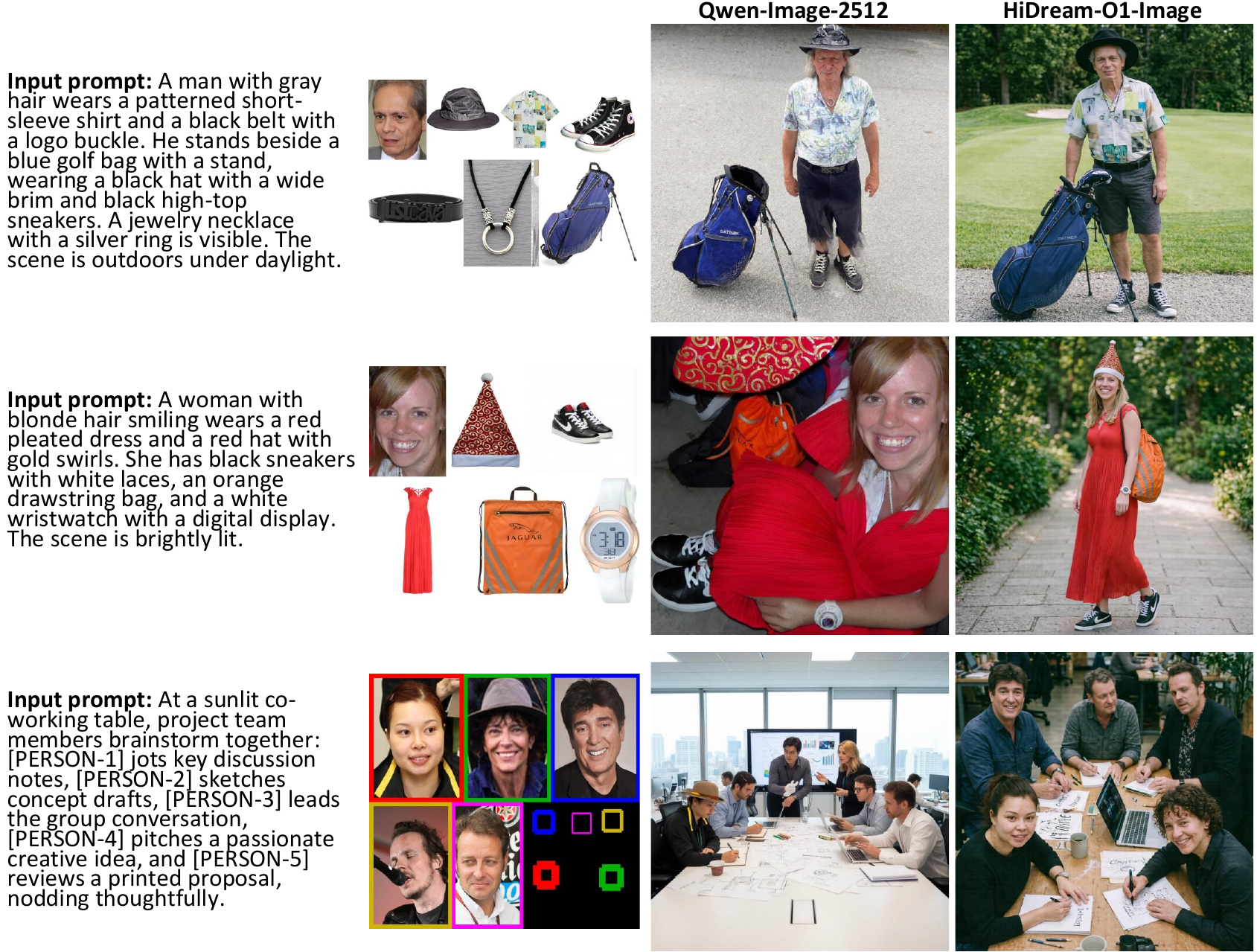} 
    \caption{Comparison between Qwen-Image and HiDream-O1-Image on subject-driven customized generation task.}
    \label{fig:subject}
        \vspace{-0.2in}
  \end{figure}

Table~\ref{tab:unisubj} details the quantitative evaluations on UniSubject. As shown in this table, our HiDream-O1-Image consistently achieves strong performances across configurations with varying numbers of reference subjects. Specifically, Scone leverages two collaborative experts via late fusion: an understanding expert provides semantic guidance to a generation expert, enabling faithful subject preservation. Echo-4o further boosts up performances by distilling knowledge from the advanced closed-source model GPT-4o to tackle the multi-reference blind spot. Compared to these baselines, our HiDream-O1-Image (8B) exhibits substantial gains via the proposed structural unification, boosting Q-O from 7.19 to 7.50 for 4--8 subjects and from 6.73 to 7.48 for 9--11 subjects. Moreover, our flagship 200B+ model pushes these boundaries even further, demonstrating superior multi-subject compositionality and identity preservation even in extreme scenarios. The core catalyst for this performance leap across both scales is our shared token space, which intrinsically bridges the feature representations of language and vision. Within this unified space, semantic concepts from textual prompts are precisely anchored to their corresponding fine-grained visual prompts (the mapped visual tokens), yielding a cohesive fused subject representation. As the number of reference subjects increases, such representation effectively mitigates interference between instruction and reference images inherent in typical disjoint encoder designs. This allows our HiDream-O1-Image to maintain consistent performance boosts under the challenging 4--8 and 9--11 subject settings.
Figure \ref{fig:subject} further showcases three subject-driven personalization results by our HiDream-O1-Image, which preserve multiple subjects better than Qwen-Image.

\section{Conclusions}
\label{sec:conclusion}
In this report, we introduced HiDream-O1-Image, a natively unified generative model with pixel-level Unified Transformer that transcends the limitations of typical disjoint latent-space VAE compression and text encoding paradigms. By dismantling the boundaries between VAEs and segregated text encoders, HiDream-O1-Image maps raw pixels, text, and task conditions into a single shared token space. This structural unification enables a consistent in-context generation process, transforming the model from a specialized T2I generator into a versatile generalist framework for diverse synthesis tasks. Extensive experiments across multiple benchmarks (e.g., GenEval, CVTG-2K, GEdit, and UniSubject) confirm the effectiveness of our framework across different scales. Notably, our highly efficient 8B model achieves competitive or superior performances in various generation tasks against state-of-the-art latent-space DiTs with significantly heavier parameters, while our scaled-up 200B+ model further pushes the boundaries of visual synthesis to establish new state-of-the-art records.

\bibliography{main}

\newpage
\appendix

\section*{Appendix}

\section{Contributions and Acknowledgments}

\definecolor{damaiblue}{RGB}{0, 0, 100}
\definecolor{damaigreen}{RGB}{0, 100, 0}
\definecolor{damaired}{RGB}{100, 0, 0}

Contributors are listed alphabetically by the last name: 

\begin{itemize}
    \item     
\textbf{Core Contributors: 
Qi Cai, Jingwen Chen, Chengmin Gao, Zijian Gong, Yehao Li, Tao Mei, Yingwei Pan, Yi Peng, Zhaofan Qiu, Ting Yao, Kai Yu, Yiheng Zhang}
\item  
\textbf{Contributors: Hao Ai, Siying Bai, Yang Chen, Zhihui Chen, Fengbin Gao, Ying Guo, Dong Li, Zhen Shen, Leilei Shi, Jing Wang, Siyu Wang, Yimeng Wang, Rui Zheng}
\item 
\textbf{Corresponding Authors: Ting Yao (tiyao@hidream.ai) and Tao Mei (tmei@hidream.ai)}
\end{itemize}

\setcounter{figure}{0}
\makeatletter 
\renewcommand{\thefigure}{A\@arabic\c@figure}
\makeatother

\setcounter{table}{0}
\makeatletter 
\renewcommand{\thetable}{A\@arabic\c@table}
\makeatother

\end{document}